\documentclass[conference]{IEEEtran}
\IEEEoverridecommandlockouts
\usepackage{cite}
\usepackage{amsmath,amssymb,amsfonts}
\usepackage[linesnumbered,commentsnumbered,ruled,lined]{algorithm2e}
\usepackage{tabularx}
\usepackage{array}   
\newcolumntype{Y}{>{\centering\arraybackslash}X}
\usepackage{graphicx}
\usepackage{subfigure}
\graphicspath{ {./imgs/} }
\usepackage{url}
\usepackage{booktabs}
\usepackage{textcomp}
\usepackage[table,xcdraw]{xcolor}
\usepackage{orcidlink}

\def\BibTeX{{\rm B\kern-.05em{\sc i\kern-.025em b}\kern-.08em
    T\kern-.1667em\lower.7ex\hbox{E}\kern-.125emX}}
    
\begin{document}

\title{An ExplainableFair Framework for Prediction of Substance Use Disorder Treatment Completion*\\
{\footnotesize \textsuperscript{*}Note: This manuscript has been accepted to the IEEE International Conference on Healthcare Informatics (IEEE ICHI 2024).}}

\author{\IEEEauthorblockN{Mary M. Lucas\orcidlink{0000-0002-0413-7499}\IEEEauthorrefmark{1}, Xiaoyang Wang\orcidlink{0000-0002-8471-4670}\IEEEauthorrefmark{1}, Chia-Hsuan Chang\orcidlink{0000-0001-9116-8244}\IEEEauthorrefmark{1}, Christopher C. Yang\orcidlink{0000-0001-5463-6926}\IEEEauthorrefmark{1},\\ Jacqueline E. Braughton\orcidlink{0000-0002-9893-2564}\IEEEauthorrefmark{2}, and Quyen M. Ngo\orcidlink{0000-0002-9781-4291}\IEEEauthorrefmark{2}}\\
\IEEEauthorblockA{\IEEEauthorrefmark{1}College of Computing and Informatics\\
Drexel University,
Philadelphia PA, USA\\ Email: mml367,xw388,cc3859,chris.yang@drexel.edu}\\

\IEEEauthorblockA{\IEEEauthorrefmark{2}Butler Center for Research
\\ Hazelden Betty Ford Foundation, Minnesota, USA\\ Email: jbraughton,QNgo@hazeldenbettyford.org}}

\maketitle

\begin{abstract}
Fairness of machine learning models in healthcare has drawn increasing attention from clinicians, researchers, and even at the highest level of government.  On the other hand, the importance of developing and deploying interpretable or explainable models has been demonstrated, and is essential to increasing the trustworthiness and likelihood of adoption of these models.  The objective of this study was to develop and implement a framework for addressing both these issues - fairness and explainability. We propose an explainable fairness framework, first developing a model with optimized performance, and then using an in-processing approach to mitigate model biases relative to the sensitive attributes of race and sex.  We then explore and visualize explanations of the model changes that lead to the fairness enhancement process through exploring the changes in importance of features.  Our resulting-fairness enhanced models retain high sensitivity with improved fairness and explanations of the fairness-enhancement that may provide helpful insights for healthcare providers to guide clinical decision-making and resource allocation. 
\end{abstract}


\section{Introduction}
Fairness of artificial intelligence (AI Fairness) has become increasingly important, drawing attention even from the highest levels of government~\footnote{https://www.whitehouse.gov/briefing-room/statements-releases/2023/10/30/fact-sheet-president-biden-issues-executive-order-on-safe-secure-and-trustworthy-artificial-intelligence/}. Developing fair models and implementing strategies to mitigate AI biases is an active ongoing field of research.  In predictive modeling for health applications, the importance of fairness goes beyond legal and ethical concerns, having significant implications for population health and the imperative to eliminate health disparities \cite{noauthor_health_nodate, wadhera_us_2023}. Many studies have evaluated bias in predictive models and attempted to improve fairness through different preprocessing, in-processing, and post-processing approaches.  Preprocessing approaches deal with the data before it goes into the model, rebalancing or reweighting it to remove disparities and imbalances that could inform the predictions and introduce bias in the model outputs.  Postprocessing approaches involve adjusting the predictions after they come out of the model.  In-processing approaches, on the other hand, involve introducing changes to the model training process itself, and are sensitive to the characteristics of the algorithm used for training. They are therefore less transparent and require more efforts to explain. Along with concerns regarding AI fairness, explainability has emerged as critical consideration when developing and implementing predictive models in healthcare \cite{yang_explainable_2022}. This focus on explainable AI must extend to fairness, allowing clinicians and others who implement these models in practice to understand what the changes are from a biased model to a fair one.

A well performing but biased model presents the risk of introducing or exacerbating health disparities, leading to poorer health outcomes or higher health costs for one demographic group over another.  A well performing model that is not explainable (``black box") may find low adoption by health care practitioners as they cannot observe and verify the rationale for the model's outputs or its decision-making process.  Bias mitigation, the process by which a model is made more fair, may often involve trading off some performance to increase fairness \cite{menon_cost_2018}.  A model that improves the fairness of the predictions across demographic groups but does not provide insight into how that fairness is achieved is equally problematic, as the lack of insight into the exact nature, extent, and impact of these trade-offs may present a challenge to the clinician as to how a patient with particular features is affected by this process. Pierson (2024) cautions that, in the context of AI fairness for health equity, ``applying quick technical fixes... without understanding what they do or whether it's relevant"~\cite{pierson_accuracy_2024} may cause more harm than good.
 
This study aims to explore the explainability of fairness enhancement. We propose ExplainableFair, a novel framework for sequentially executing fair model training and providing explanations for enhanced fairness. Under this framework, we first develop a model with well-tuned predictive performance. We then utilize an in-processing approach on the trained model for bias mitigation. Finally, we employ feature importance analysis to interpret fairness improvement, applying clarifying questions to explain why the model becomes more fair by examining how the importance of different features varies across the fairness enhancement process. We apply our framework to the task of predicting failure to complete substance use disorder (SUD) treatment.

\section{Literature Review}
In this section we review relevant studies related to predictive modeling in SUD, AI fairness, and explainability. 

\subsection{Predictive Modeling}
Ethical concerns over the use of AI and ML in different domains have emerged over the past few years, with bias identified in several commercial AI applications, one of the most prominent being a model used in criminal recidivism risk prediction that demonstrated bias against Blacks~\cite{angwin_machine_nodate,chouldechova_fair_2017}. In the healthcare domain, an algorithm used to identify and help patients with complex health needs exhibited bias against Black patients, leading to reduced resources being allocated to these patients~\cite{obermeyer_dissecting_2019}.
Researchers have used ML approaches in SUD treatment for various tasks, for example predicting outcomes of SUD treatment~\cite{acion_use_2017,nasir_machine_2021}, and specifically readmission or relapse after SUD treatment~\cite{liang_developing_2021,morel_predicting_2020}. While these studies individually and collectively explore multiple ML approaches and algorithms, identify important predictors, and report good performance, they do not report any bias evaluation or how the models perform for different demographic groups. Our previous work used preprocessing by resampling to improve fairness of models predicting SUD treatment completion failure~\cite{lucas_resampling_2023}, but did not extend into explaining the fairness enhancement beyond group distribution imbalances.

\subsection{Bias Mitigation through In-processing}

In-processing approaches for bias mitigation learn a fair model by modifying the model learning process. Learning fair representation~\cite{zemel_learning_2013,zhang_mitigating_2018,adel_one-network_2019} is one of two main approaches which is focused on learning a generative model that maps each data point into an arbitrary representation. The goal of the generative model is to generate a representation that reserves the ability to predict the target label while decoupling the dependence on sensitive attributes. The other main in-processing approach is regularized optimization~\cite{kamishima_fairness-aware_2012,agarwal_reductions_2018,roh_fairbatch_2021,shen_optimising_2022}, which learns a predictive model by adding a disparity regularization term with the performance loss. The common regularization terms are group-based fairness metrics, such as demographic parity, equalized opportunity, and equalized odds. Since most of in-processing bias mitigation approaches focus on reaching a fair model, they lack in reporting of feature importance change during the fairness optimization.

\subsection{Explainable AI in Healthcare}
The importance of explainability for creating trust in AI-informed decision making has been discussed in many studies.  Angerschmid et. al. (2022) examine the effects of fairness and explanations through a case study, using example-based explanations and feature importance-based explanations \cite{angerschmid_effects_2022}. Their study reveals that decisions accompanied by explanations result in increased trust. However, introducing fairness information has mixed results that reflects the complexity of deploying AI explanation and fairness statements.  Yang, C. (2022) \cite{yang_explainable_2022} discusses the importance of explainability in health AI  and frames the process through two types of questions - explanation and clarification.  The explanation questions are information-based whereas the clarification questions are instance-based, where the required information is generated as the model is executed to provide additional information regarding the predictions. Combi et. al. (2023)~\cite{combi_ihi_2023} highlight the importance of tailoring ``both the way and the concepts used in explanations" to ensure they are fitting for the end user or stakeholder. Zhou, Chen, and Holzinger (2020) \cite{zhou_towards_2022} provide an overview of the relationship between AI fairness and explanation and conclude that fairness requires ``comprehensive contextual understanding" and that AI explanations can contribute to this.   

\section{Materials and Methods}
\subsection{The data}
The data used for this study was extracted from the Hazelden Betty Ford Foundation (HBFF) electronic health record (EHR).  HBFF is one of the largest nonprofit addiction treatment providers in the United States. HBFF data contain information on 20 years of patient encounters from 2000 to 2019 and is described in more detail in~\cite{liang_developing_2021}. There are multiple variables available, including demographic information (e.g., race, ethnicity, age, legal sex), other patient related and socioeconomic variables (e.g., education level, employment status, occupation, marital status), encounter-specific variables (e.g. length of stay, primary diagnosis, discharge status), diagnosis-related variables (e.g., substance used, co-occurring mental health diagnoses), and variables encoding responses to clinical questionnaires.
A distinct feature of this dataset is that unlike most structured EHR data that typically comprise objective clinical measurement variables (e.g. vital signs, laboratory measurements, medications administered, etc.), this dataset also contains multiple fields comprising answers to questionnaires administered to the patient during their stay and treatment at HBFF treatment facilities.  These include the American Society of Addiction Medicine (ASAM) Criteria which measure substance use severity and are grouped into six dimensions \cite{liang_developing_2021}.  Also included is information on the types of services the patient utilized in their treatment journey. 

We included inpatient and outpatient encounters, using the discharge status variable to determine treatment completion.  To reduce uncertainty in the discharge status we excluded encounters where patients had a transfer, and only included encounters with discharge ``with staff aproval'' (WSA), ``conditional with staff approval'' (CWSA), ``against staff/medical advice'' (ASA/AMA), or ``at staff request'' (ASR).  
Encounters with the discharge status of ASA/AMA and ASR were considered to not have successfully completed treatment. Since the aim is to build a model to predict failure to complete treatment, these encounters were labelled as positive (class label 1) and those with status WSA or CWSA were labeled as having completed treatment (class label 0).  
Additional data preparation included aggregating race groups with small sample sizes and categorising race as ``Caucasian'' or ``Not Caucasian''. The final dataset comprised 10,673 encounters for 9,369 distinct patients, and is highly imbalanced both with respect to the target variable and the demographic group distributions. Table~\ref{tab:dataset-characteristic} shows the group distributions stratified by treatment completion status. 

\begin{table}[ht]
\centering
\caption{Distribution of patients by sensitive attributes and class label}
\label{tab:dataset-characteristic}
\resizebox{\columnwidth}{!}{%
\begin{tabular}{@{}lllll@{}}
\toprule
Characteristic                        & Negative Class (9,149) & Positive Class (1,524) \\ \midrule
Race                     &              &                          \\
\text{    }Caucasian               & 8,230 (90\%) & 1,341 (88\%)                   \\
\text{    }Not Caucasian          & 919 (10\%)   & 183 (12\%)                     \\ 
Sex                      &              &               \\
 \text{    }Male                     & 5,824 (64\%) & 1,062 (70\%)                   \\
\text{    }Female                   & 3,325 (36\%) & 462 (30\%)                     \\\bottomrule
\end{tabular}%
}
\end{table}

To obtain more reliable and robust assessments, we divided the dataset into a training set $D_{train}$ (90\%) and a test set $D_{test}$ (10\%) ten times, with random different train-test splits on each occasion. The distribution of sensitive attributes and the target variable is similar across each combination of training and test sets. 

\subsection{ExplainableFair Framework}
\begin{figure*}[t]
    \centering
    \includegraphics[width=0.9\linewidth]{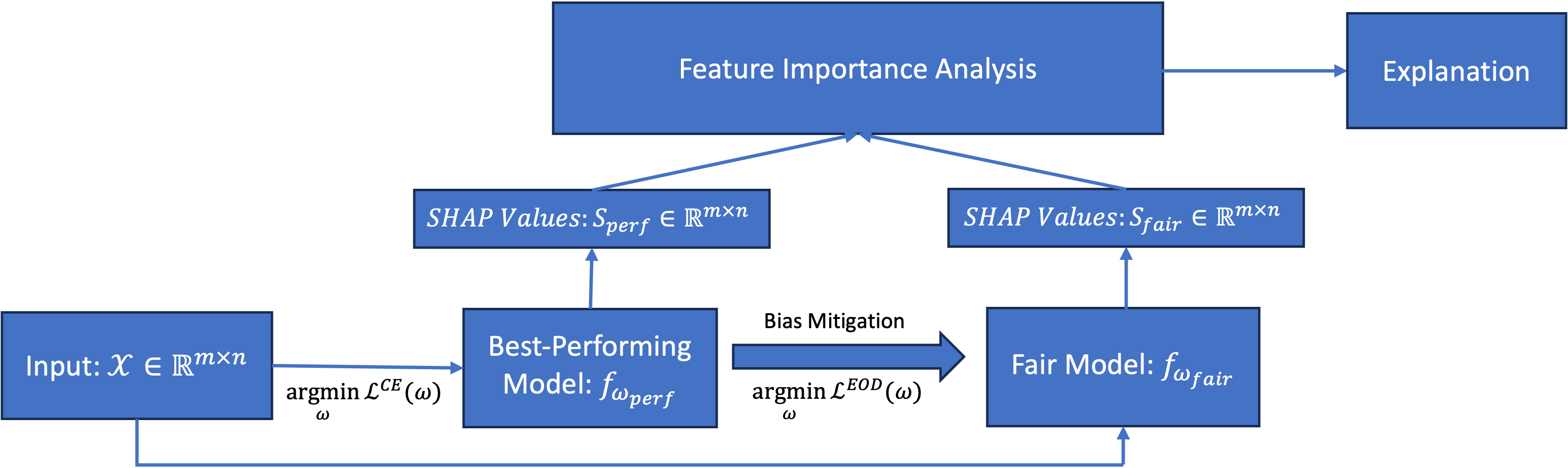}
    \caption{The ExplainableFair framework.}
    \label{framework}
\end{figure*}

As delineated in Figure~\ref{framework}, the ExplainableFair framework is bifurcated into two distinct phases: the model training phase and the fairness explanation phase. During the model training phase, we train a Logistic Regression model \(f_{\omega_{perf}}\) to maximize predictive performance. Subsequently, this trained model undergoes a fine-tuning process for addressing fairness concerns, resulting in a fair model \(f_{\omega_{fair}}\) with equitable performance across demographic groups related to sensitive attributes. The fairness explanation phase then builds upon \(f_{\omega_{perf}}\) and \(f_{\omega_{fair}}\). Utilizing the SHapley Additive exPlanations (SHAP) values~\cite{lundberg_unified_2017} obtained from the two models, we analyze the changes in feature importance attributable to the fairness optimization. This involves a comparative assessment of the most important features and a focused examination of those features that exhibit the most pronounced changes between two models. The implementation of this framework not only enhances the fairness of the resulting model but also bolsters its interpretability, particularly for clinical contexts.

\subsection{Predictive Modeling}

We select the logistic regression model as our classifier due to its robustness and interpretability. Given a patient $x \in R^m$ with the target variable $y$, where $m$ is the number of features, the learning process of the logistic regression model is to find the coefficient value $\omega \in R^m$ that reduces the difference between actual value $y$ and the predicted value $f_\omega(x) = \hat{y}$. Because of the class imbalance in our dataset, we set the classification threshold using ``The Closest to (0,1) Criteria (ER)", which uses the point minimizing the Euclidean distance between the ROC curve and the (0,1) point~\cite{unal_defining_2017,perkins_inconsistency_2006}. Therefore, we can find the optimal model $f_{\omega_{perf}}$ that minimizes the entropy loss in all patients~$X$, where the coefficient $\omega_{perf}$ is identified by: 

\begin{equation}
\label{ce}
\text{argmin}_{\omega} \ \mathcal{L}^{CE}(\omega) = -E_{x \sim X}\left( y \log(\hat{y}) + (1 - y) \log(1 - \hat{y}) \right)
\end{equation}

\subsection{Bias Mitigation}

In this stage, we fine-tune $f_{\omega_{perf}}$ to optimize the model fairness. We select \emph{Equalized Odds (EO)}, a comprehensive measure described by Hardt et al.~\cite{hardt_equality_2016}, as the fairness criterion. EO mandates that both the True Positive Rate (TPR) and False Positive Rate (FPR) exhibit minimal disparity across different demographic groups. Based on this concept, the proposed loss function incorporates the \emph{Equalized Odds Disparity (EOD)}, denoted as \(\mathcal{L}^{EOD}\), which is the sum of the differences in TPR and FPR as presented in Equation~\ref{EOD_loss}: 

\begin{equation}
    \mathcal{L}^{EOD}(\omega) = \mathcal{L}_{TPR} + \mathcal{L}_{FPR},
    \label{EOD_loss}
\end{equation}

\noindent where the least square error is used for the calculation of TPR (FPR) difference loss term: 
\begin{align}
    \mathcal{L}_{TPR} &= (TPR_{z=0} - TPR_{z=1})^2 \\
    \mathcal{L}_{FPR} &= (FPR_{z=0} - FPR_{z=1})^2 \\
    TPR_{z} &= Pr(\hat{y}=1 |z=z, y=1 ) \\
    FPR_{z} &= Pr(\hat{y}=1 |z=z, y=0 )
\end{align}

Note that \(z \in \mathcal{Z}\) is defined as a sensitive attribute. Therefore, for example, when optimizing the pre-learned model for a race-fair model, we set $z=1$ for Caucasian and $z=0$ for Non-Caucasian. Similarly, to achieve a sex-fair model, we set $z=1$ for males and $z=0$ for females. This \(\mathcal{L}^{EOD}\) loss term ensures that the TPR and FPR differences between demographic groups are minimized simultaneously. As a result, we use $\mathcal{L}^{EOD}$ to keep optimizing the coefficient values of $f_{\omega_{perf}}$ and obtain a fair model $f_{\omega_{fair}}$ after the optimization.

\subsection{Feature Importance Analysis}

We utilize a local feature attribution method, SHAP, which distributes the prediction score of a fitted model for a patient~$x \in X$ to its base features~$R^{m}$. The score of a base feature can be interpreted as the importance of the feature to the patient. With this method, we propose Algorithm~\ref{alg:feature-importance} to analyze how the feature importance vary between $f_{\omega_{perf}}$ and $f_{\omega_{fair}}$. In the Algorithm, line 1 and 2 initialize two SHAP explainers using DeepSHAP~\cite{lundberg_unified_2017} . Since the SHAP value is model-dependent, we initialize $E_{perf}$ and $E_{fair}$ for $f_{\omega_{perf}}$ and $f_{\omega_{fair}}$, respectively. We then apply the explainers to calculate the SHAP values for the test dataset (line 3), which result in two feature importance matrices $S_{perf}, S_{fair} \in R^{n \times m}$. $n$ is the number of patients in the testing dataset, and $m$ is the number of features. Each row represents a patient with $m$ importance scores, where each corresponds to a feature. 

\begin{algorithm}[htbp]
\label{alg:feature-importance}
\caption{Feature Importance Analysis using SHAP Values}
\SetKwInOut{Input}{Input}\SetKwInOut{Output}{Output}
\SetKwComment{comment}{\#}{}
\Input{Model before Bias Mitigation $f_{\omega_{perf}}$, Fair Model after Bias Mitigation $f_{\omega_{fair}}$, Train set $X_{train}$, Test set $X_{test}$
}
\Output{Feature importance rankings $R_{perf}, R_{fair}$, Most Changed features Set $C$}
\BlankLine

$E_{perf} \leftarrow shap.DeepExplainer(f_{\omega_{perf}}, X_{train})$\;
$E_{fair} \leftarrow shap.DeepExplainer(f_{\omega_{fair}}, X_{train})$\;
$S_{perf}, S_{fair} \leftarrow E_{perf}(X_{test}), E_{fair}(X_{test})$\;
$\bar{S}_{perf}, \bar{S}_{fair} = mean(S_{perf}), mean(S_{fair})$\;

$R_{perf} \leftarrow RankFeatures(\bar{S}_{perf})$\;
$R_{fair} \leftarrow RankFeatures(\bar{S}_{fair})$\;
$C \leftarrow CompareRankings(R_{perf}, R_{fair})$\;
\end{algorithm}

To quantify the the extent of importance changes for each feature across the best-performing model and the fair model, we propose to apply two operations:

\begin{enumerate}
    \item Aggregation: as shown in line 4, we aggregate feature importance scores of each feature by taking average (i.e., mean function in the algorithm) across $n$ patients, which results in $\bar{S}_{perf}, \bar{S}_{fair} \in R^{1 \times m}$.
    \item Rank transformation: SHAP values are model-dependent, so it is invalid to compare the aggregated importance scores between models. To overcome this, we sort features in $\bar{S}_{perf}$ and $\bar{S}_{fair}$ in a descending order and use the rank to represent the importance of each feature (line 5 and 6). $R_{perf}$ and $R_{fair}$ are the outputs, and they denote the importance ranks of features.
\end{enumerate}

Consequently, we can use difference of rank between $R_{perf}$ and $R_{fair}$ to identify the features whose ranking changes the most. This helps us to highlight key features that may be critical for fairness, potentially providing the insights for practitioners. Moreover, to provide the robust analysis, we apply Algorithm~\ref{alg:feature-importance} for ten different training and testing splits. Our experimental results are the aggregated results of ten splits.

\subsection{Performance and Fairness Evaluation}
Predictive performance was evaluated using Area Under Receiver Operating Characteristic Curve (AUROC), sensitivity, and specificity. In the context of our study, we operationalized the concept of the `privileged group' based on dataset representation with respect to a sensitive attribute. Specifically, the privileged group for a specific sensitive attribute is defined as the group having a larger representation within the dataset. To quantify the fairness of the model, we employed the EOD, mathematically defined as the arithmetic mean of the differences in the TPR and FPR across privileged groups and unprivileged groups, formulated as follows:
\begin{equation}
EOD = \frac{\mathcal{L}_{TPR} + \mathcal{L}_{FPR}}{2}
\label{eod_defn}
\end{equation}
To ensure the robustness and reliability of our findings, each model training iteration was repeated ten times. We then computed and reported the average values across these iterations for key metrics, including predictive performance, fairness levels, and feature importance.

\section{Results}

In Tables~\ref{tab:performances_and_fairness_race} and \ref{tab:performances_and_fairness_sex}, we compare the model performance before and after fairness optimization for each sensitive attribute (race and sex). 

\begin{table}
\centering
\caption{Model performance and fairness - Race-Fair Optimization}
\label{tab:performances_and_fairness_race}
\begin{tabularx}{\columnwidth}{lYYYY}
\hline
                      & \centering{AUROC}  & \centering{Sensitivity} & \centering{Specificity} & \centering{EOD}\tabularnewline
\hline
Best Performing Model & 0.8607 & 0.7948      & 0.8054      & 0.0725 \\ 
Race-Fair Model       & 0.8585 & 0.8037      & 0.7533      & 0.0298 \\ 
\hline
\end{tabularx}
\end{table}

\begin{table}
\centering
\caption{Model performance and fairness - Sex-Fair Optimization}
\label{tab:performances_and_fairness_sex}
\begin{tabularx}{\columnwidth}{lYYYY}
\hline
                      & \centering{AUROC}  & \centering{Sensitivity} & \centering{Specificity} & \centering{EOD}\tabularnewline
\hline
Best Performing Model & 0.8607 & 0.7948      & 0.8054      & 0.0603 \\ 
Sex-Fair Model        & 0.8576 & 0.7829      & 0.7545      & 0.0282 \\ \hline
\end{tabularx}
\end{table}

\begin{figure}[ht]
    \centering
    \includegraphics[width=1\linewidth]{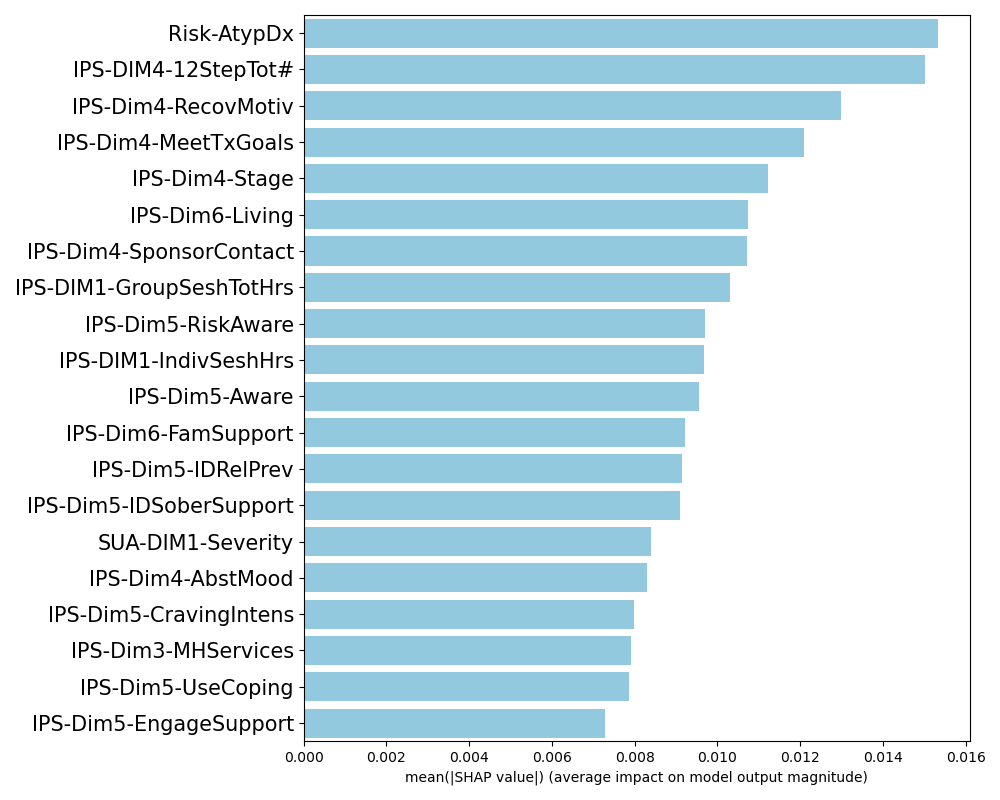}
    \caption{Most important features before fairness optimization.}
    \label{best-most}
\end{figure}

\begin{figure}[ht]
    \centering
    \includegraphics[width=1\linewidth]{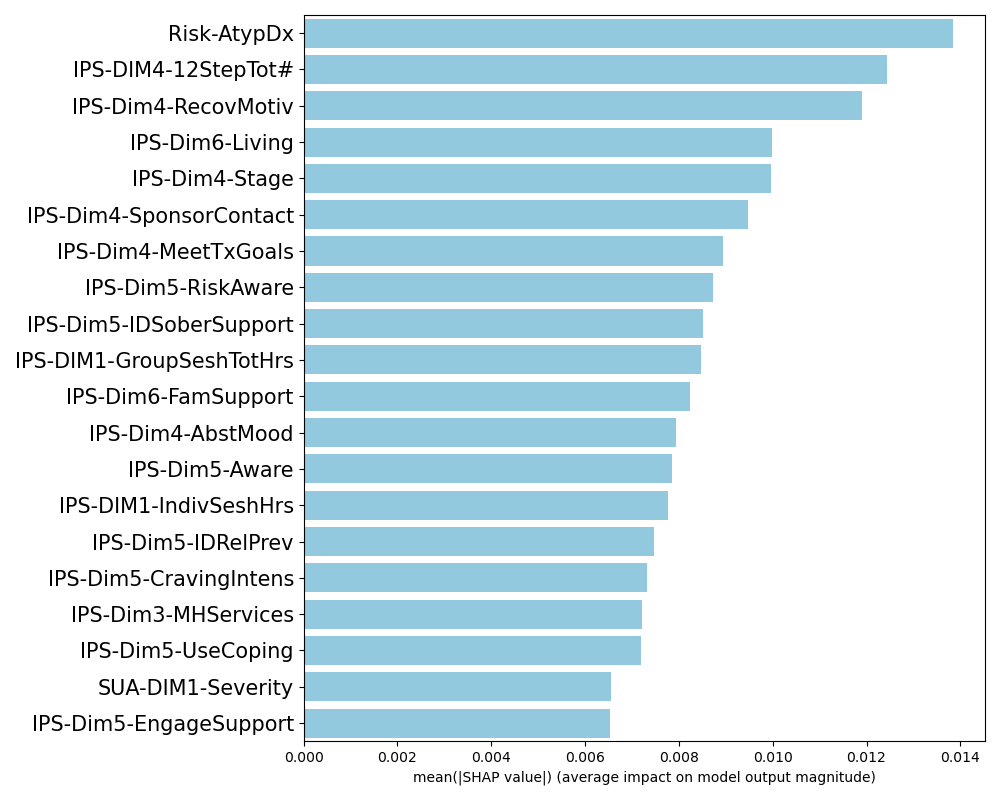}
    \caption{Most important features after \textit{race-fair} optimization.}
    \label{race-most}
\end{figure}

\begin{figure}[ht]
    \centering
    \includegraphics[width=1\linewidth]{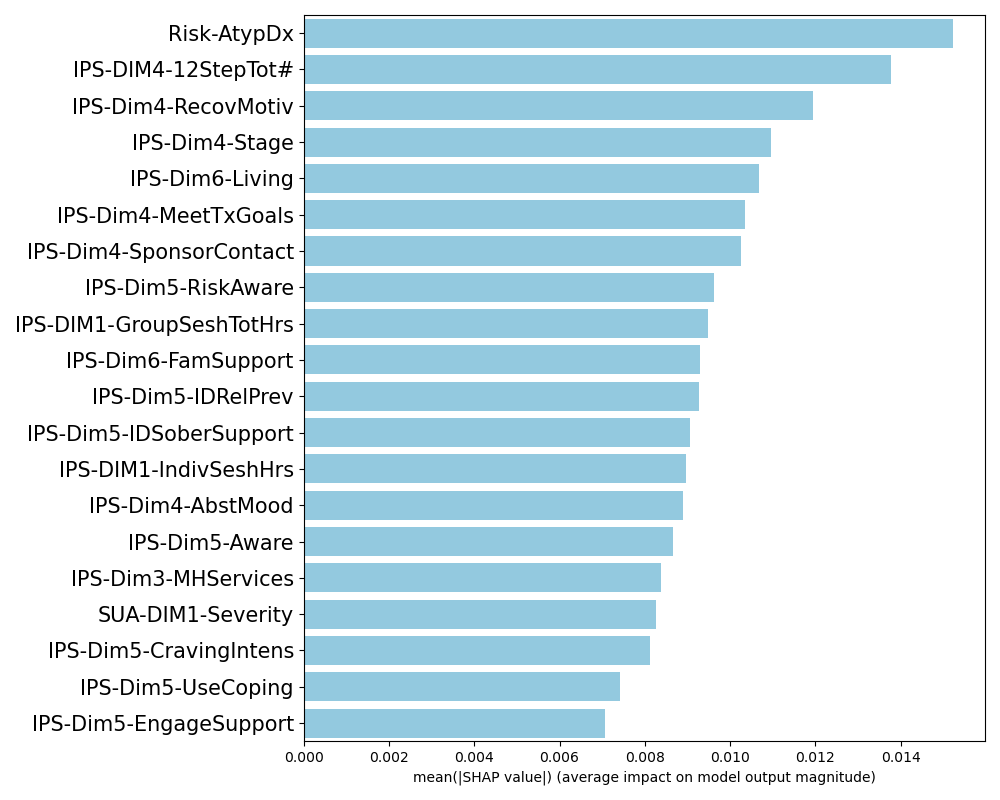}
    \caption{Most important features after \textit{sex-fair} optimization.}
    \label{sex-most}
\end{figure}

Figures~\ref{best-most} through \ref{sex-most} present the twenty most important features, based on SHAP, for the best performing model before fairness optimization, and for the race-fair model and sex-fair model respectively.  We discuss the most important of these in the Discussion section.

\begin{figure}[ht]
    \centering
    \includegraphics[width=1\linewidth]{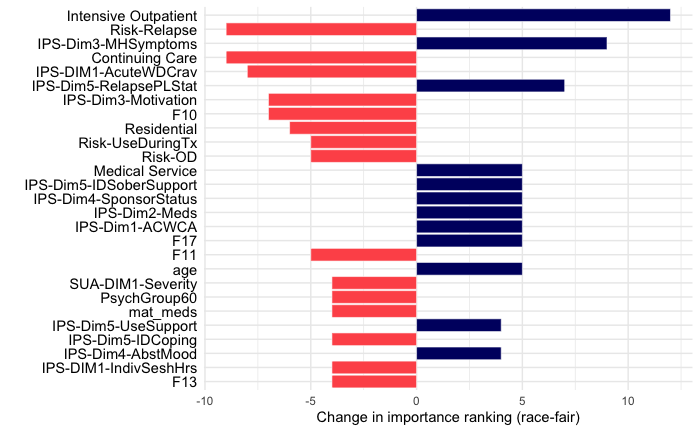}
    \caption{Most changed features (by ranking) during race fairness optimization}
    \label{race-changed-most}
\end{figure}

\begin{figure}[ht]
    \centering
    \includegraphics[width=1\linewidth]{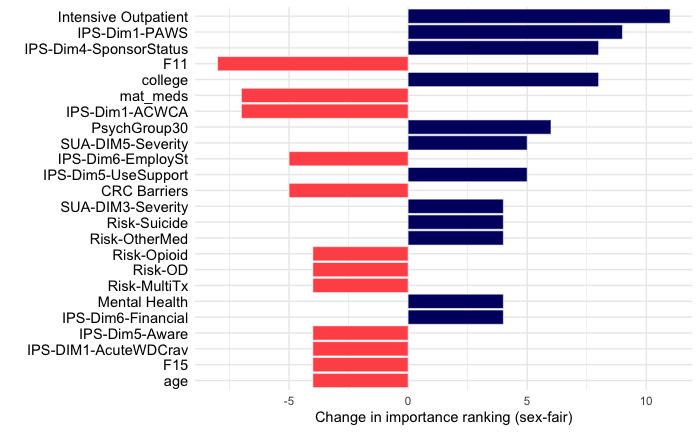}
    \caption{Most changed features (by ranking) during sex fairness optimization}
    \label{sex-changed-most}
\end{figure}

In figures~\ref{race-changed-most} and \ref{sex-changed-most} we show the features whose importance ranking changes the most during race-fairness and sex-fairness optimization respectively. The bars to the right (in blue) indicate increase in importance ranking while those to the left (red) indicate decrease in ranking. We again expand on the most important of these, with respect to change in ranking, in the Discussion section.

\section{Discussion}
The overall purpose of healthcare predictive modeling is inherently pragmatic, applying data-driven results into real-world implications for clinical providers and patients. In the field of alcohol and substance use treatment and recovery, these implications can have life-altering ramifications for patients’ long-term recovery. There is a wide, and ever-growing gap in accessibility and quality of treatment for those with substance use disorders who need, seek, initiate, and attend treatment.  This is particularly true for those from marginalized groups (e.g., racial minority, gender/sexual minority).

As use of machine learning and AI continues to grow in driving healthcare decisions, model features associated with optimal treatment outcomes, with their potential underlying biases, will directly impact the treatment and access that may be available to patients. 

In this study, we introduced a novel framework that develops a fair model for predicting SUD treatment completion using in-processing methods, and explained the enhancement of fairness through feature importance analysis. We adapted the explainability structure of questions in \cite{yang_explainable_2022}, and crafted fairness-relevant explanation and clarification questions to guide our work.  Because we are concerned with fairness across demographic groups, we framed the clarification questions as group-based rather than instance-based (Table~\ref{tab:explanation_qns}).

\begin{table*}[htbp]

\caption{Information-based and Clarification-based questions for explaining model fairness optimization}
\label{tab:explanation_qns}
\resizebox{\textwidth}{!}{%
\begin{tabular}{@{}ll@{}}
\toprule
\multicolumn{2}{l}{Information-based explanation questions}                                                                                                                                                                                                                                                                                                                                                             \\ \midrule
Best performing model    & \begin{tabular}[c]{@{}l@{}}For the best performing model before fairness optimization, what is the performance for the whole population?\\ \\ What are the most important features? \\ \midrule \end{tabular}            \\ 

Fairness-optimized model & \begin{tabular}[c]{@{}l@{}}After optimizing the fairness while retaining performance, what is the performance for the whole population?\\ \\ What are the most important features?\end{tabular}                                                                                                                                                                   \\ \toprule
\multicolumn{2}{l}{Group-based clarification questions}                                                                                                                                                                                                                                                                                                                                                                 \\ \midrule
What if ?                & \begin{tabular}[c]{@{}l@{}}What are the differences between groups in the best performing model?\\ \\ What are the differences between groups in the fairness-optimized model?\\ \\ What are the fairness measures used? \\ \\ What is the improvement in fairness? \\ \\ What is the sacrifice in performance to achieve fairness? \\ \midrule \end{tabular}                                                                                            \\ 
                         & \begin{tabular}[c]{@{}l@{}} \\ \midrule \end{tabular} \\ 
How to be that?          & \begin{tabular}[c]{@{}l@{}}Does the importance of features change after fairness optimization?\\ \\ What features become more important when the fairness is optimized based on a selected fairness criteria or measure?\\ \\ What features become less important when the fairness is optimized based on a selected fairness criteria or measure?\end{tabular}                          \\ \bottomrule
\end{tabular}%
}
\end{table*}

We used logistic regression to train a best performing model, and then optimized it for fairness as described in the Methods section. Using SHAP values to determine the feature importance ranking of the different features in the base and fairness optimized models, we addressed the fairness-relevant questions to develop an understanding and explanation of the fairness enhancement process.  

\subsection{Information-based explanation questions}
The results of model performance as measured by AUROC, sensitivity, and specificity,  presented in Tables~\ref{tab:performances_and_fairness_race} and \ref{tab:performances_and_fairness_sex} guide us to answer the information-based questions regarding the performance of the best model before and after fairness optimization. The best performing model has an AUROC of 86.07\%, sensitivity of 79.48\% and specificity of 80.54\%.  The race-fair and sex-fair model have AUROC 85.85\% and 85.76\%, sensitivity 80.37\% and 78.29\%, and specificity 75.33\% and 75.45\% respectively.  Thus we observe that both the best performing and the fairness optimized models have acceptable performance, above 75\%, on all metrics.  

We obtain insight into the most important features before and after fairness optimization from figures~\ref{best-most} to \ref{sex-most}.  We observe that the most important features for predicting failure to complete SUD treatment remained fairly stable across the fairness optimization process.  In the best performing model, the most predictive features are documented risk of atypical discharge (\textit{``Risk-AtypDx"}) and scores on the various dimensions of the ASAM Criteria. Of these, the most dominant are those from dimension 4 of the criteria, which assess the readiness to change, specifically participation in the 12-Step program (\textit{``IPS-DIM4-12StepTot\#"}), motivation to recover (\textit{``IPS-Dim4-RecovMotiv"}), meeting treatment goals (\textit{``IPS-Dim4-MeetTxGoals"}), and stage of recovery process (\textit{``IPS-DIM4-Stage"}). These five most important features remain stable through the fairness optimization, although we do observe from figures~\ref{race-most} and \ref{sex-most} that the patient's living environment (\textit{``IPS-Dim6-Living"}) becomes more important in the fair models (this is an element of dimension 6 of the ASAM criteria, which assesses the patients recovery environment).

\subsection*{Clinical Implications}
The results in figure~\ref{best-most}, before fairness optimization, with exception to risk for atypical discharge, identifies themes associated with individual readiness and motivation to change in four of the top five most important factors for treatment. In this model, clinicians would likely be instructed to place emphasis in encouraging patients, regardless of their motivation and readiness to change, to find and regularly attend a local 12-step, peer recovery support group. Previous empirical findings support that consistent involvement in community groups, such as 12-step/peer recovery support groups, is a salient factor in sustained recovery~\cite{reif_peer_2014, leamy_conceptual_2011, best_recovery_2017}. In addition, clinical treatment goals and associated interventions would likely focus on guiding patients to identify their readiness and subsequent motivation for change (i.e., recovery), and then exploring ways to build upon, maintain, or challenge these beliefs. As a result, meeting individual treatment goals would likely mean that patients have shown willingness to, and actionable change towards, developing skills that increase their confidence/motivation of sustaining change and building a solid recovery support system. 

Comparison between figures~\ref{best-most} and \ref{race-most}, as well as figures~\ref{best-most} and \ref{sex-most} illustrate similarities across the top three features even after optimizing for race-fairness and sex-fairness. 
Yet, the optimization for both race- and sex-fairness denotes a distinct difference, raising the priority of living conditions and stage of recovery, and decreasing the overall importance of meeting treatment goals as an indicator for successful SUD treatment for non-White patients. This difference may be because marginalized groups, including women, tend to face additional barriers to meeting treatment goals~\cite{center_for_behavioral_health_statistics_and_quality_racialethnic_2021, hall_experiences_2022, sahker_substance_2020}. Effective treatment goals are relevant and achievable; in other words, they must be relevant to the collection of symptoms/diagnosis presented and reflect consideration of individual contextual factors and resourcing. Without this consideration, treatment goals are ineffective at best, and may result in traumatization at worst. While gaining diversity, much of the extant research on substance use treatment and recovery in the United States is based on majority White, heterosexual, biological male samples. Evidence-based treatment goals are therefore, often standardized and conceptualized through the experience of the majority (i.e., in this case, White, biological male, westernized medical model) patients~\cite{wagner_recovery_2020}. As a result, sole use of meeting treatment goals without meaningful changes that reflect consideration of historical, environmental, and social context that disproportionately affect marginalized communities (including women) may result in ineffective treatment and decreased patient outcomes for those populations. 

In the race- and sex-fair models, clinicians would likely focus interventions to helping individuals identify their stage of recovery and ensuring individuals developed the tools necessary and connected with appropriate resources to ensure their living conditions/day-to-day responsibilities were conducive to recovery.

\subsection{Group-based clarification questions}
The ``What if?" clarification questions address the overarching question ``what would happen if...?", i.e. what changes occur if we optimize the best performing model for fairness.  We refer again to tables~\ref{tab:performances_and_fairness_race} and \ref{tab:performances_and_fairness_sex} for the Equalized Odds Disparity (EOD) between race groups and sex groups respectively. This is the arithmetic mean of the difference in TPR and the difference in FPR between groups and is previously described in eqn~\ref{eod_defn}. The EOD between race groups in the best performing model is 0.0725 while that in the race-fair model is 0.0298.  In comparing male vs female groups, the EOD for the best performing model is 0.0603 while that of the sex-fair model is 0.0282.  We use the EOD as the fairness measure. 

In looking at model performance, we observe that there was a slight drop (less than 1\%) in AUROC after optimizing the model for race-fairness and for sex-fairness. 
The sex-fair model also had a 1.2\% drop in sensitivity, while both fairness optimized models had an approximately 5\% drop in specificity. We note a significant increase in fairness, with a drop in EOD of 4.27\% in the race-fair model and 3.21\% in the sex-fair model. Because the positive class in our prediction task is ``did not complete treatment", we place a higher priority on the sensitivity of the model.  This is because having a drop in TPR would be costly, resulting in missing patients who could benefit from additional interventions to help them complete treatment. Our fairness optimized models maintain high performance, close to that of the best performing model while simultaneously exhibiting less bias.

The ``How to be that?" clarification questions on the other hand may be understood to address the issue of ``how do we get there from here?" How do the features change to attain model fairness?  We answer these questions by examining figures~\ref{race-changed-most} and \ref{sex-changed-most}.  
For our logistic regression model to become race-fair the features that increase most in importance ranking include the number of days spent in intensive outpatient treatment (\textit{``Intensive Outpatient"}), presence of mental health symptoms (\textit{``IPS-Dim3-MHSymptoms"}), and the status of relapse prevention planning (\textit{``IPS-Dim5-RelapsePLStat"}). The features that drop most in importance ranking include documented risk of relapse (\textit{``Risk-Relapse"}), number of days in spent in the Continuing Care program (\textit{``Continuing Care"}), and acute withdrawal and craving (\textit{``IPS-DIM1-AcuteWDCrav"}). 
For the sex-fair model, the features that increase most in importance ranking include the number of days spent in intensive outpatient treatment (\textit{``Intensive Outpatient"}), current post-acute withdrawal syndrome (\textit{``IPS-Dim1-PAWS"}), status of sponsor relationship (\textit{``IPS-Dim4-SponsorStatus"}), and having a college education (\textit{``college"}). On the other hand, having a diagnosis related to opioid use disorder (\textit{``F11"}), having a record of medications in the EHR medication administration table (\textit{``mat\_med"}) , and scores on Acute Withdrawal Craving Answers (\textit{``IPS-Dim1-ACWCA"}) drop in importance ranking.

\subsection*{Clinical Implications}
In reviewing figure~\ref{race-changed-most}, we see several important changes. First, the decreased importance of (a) continuing care and (b) individual motivation to recover is contrary to widely-held empirically-based beliefs that both are critical elements for successful SUD treatment outcomes. These results suggest that the type of continuing care offered may not be as helpful to marginalized communities and may be more tailored for this sample’s dominant racial group. In like manner, the race-fair model showed an increased importance of intensive outpatient treatment (IOP), which allows the ability to receive treatment services while patients remain within their communities. This suggests that successful treatment and continuing care plans must include ways for patients from marginalized groups to work on their recovery within their own community, where they are much more likely to have a felt sense of belonging, likely resulting in increased feelings of safety and having more practical, sustainable supports~\cite{best_recovery_2017, wagner_recovery_2020, marino_belong_2015}. Decreased emphasis in motivation to change may also suggest that regardless of how motivated a patient is to complete treatment and recover, there may still be enduring (individual, environmental, structural, etc.) barriers that forcibly hinder an individual’s ability to meet treatment goals and sustain recovery, which is often the case for marginalized groups.  For example, integration of co-occurring mental health symptoms also increases in importance. Overall, statistics support that marginalized groups have higher incidences and severities of co-occurring medical and psychiatric symptoms due to long-standing barriers to care and ineffective treatment~\cite{center_for_behavioral_health_statistics_and_quality_racialethnic_2021, hall_experiences_2022, sahker_substance_2020}.  In comparison, it seems that factors that tend to be most important for successful SUD treatment for the dominant racial groups may not accurately account for structural barriers and complexity of cases that may be present for marginalized communities.  Overall, the features that increased in importance in the race-fair model – IOP, mental health symptoms, sober support, sponsor status – illuminate the consideration of complexity of cases, barriers to care, vital need for community/sense of belonging during SUD treatment, and recovery support for marginalized communities. 

From figure~\ref{sex-changed-most}, we also discover several important changes. For the sex-fair model, there is an increased importance in having a college education, which is not observed in the race-fair model. Previous research supports that for patients of color, socio-economic factors (e.g. education level or income) do not appear to benefit them within the healthcare system. For example, maternal mortality for Black women in the United States is high regardless of individual income level or education status. In contrast, without intersectional factors of race and ethnicity, education status tends to be significant for female/female-identifying patients as higher education typically increases their knowledge of and access to care. In addition, the sex-fair model illustrated increased importance in post-acute withdrawal syndrome (PAWS), participation within an intensive outpatient treatment program, and obtaining a sponsor/recovery mentor. As with the race-fair model, integration of flexible treatment programs, such as an intensive outpatient treatment program, may help women get the care that they need without having to abandon day-to-day responsibilities and community support. This may be especially important for women with high parental/childcare and household management responsibilities, who may otherwise not be able to seek and receive treatment. Increased emphases in obtaining sponsors illuminates a need for women to feel like they belong and have mentors/social support within the recovery community. Due to historical stigma where women, specifically mothers, struggling with substance use are seen as amoral, and much of the treatment and recovery community being comprised of male peers, women may still struggle to find sponsors that can provide relevant and meaningful support. This suggests that clinical treatment may want to include specific interventions and resources towards identifying and solidifying a recovery community in which female-identifying patients feel a sense of belonging.  While gaining more attention, PAWS is still viewed as a novel phenomenon with some evidence of gender differences. Yet, these results suggest the importance of regular assessment and integration of medical treatment for post-acute withdrawal for female/female-identifying patients. Future research should investigate the differences in PAWS prevalence among gender and race/ethnicity patient populations. As with the race-fair, the sex-fair model illustrates unique features of increased importance (e.g., college education, sponsor status, intensive outpatient program, PAWS) for ensuring successful treatment for female/female identifying patients. 

Though there are salient similarities across all models, the nuanced differences illustrated in figures~\ref{race-changed-most} and \ref{sex-changed-most} have critical ramifications for implementing elements of successful treatment experiences across diverse patient populations.

\section{Conclusion}
In this study we present ExplainableFair framework, a novel approach to developing a fair model and explaining the fairness enhancement by comparing feature importance before and after fairness optimization. 

Model unfairness can come from many sources. These include biases encoded in datasets, biases resulting from the algorithm used for training, and biases caused by the features used in the model \cite{pessach_review_2023}. It is important to quantify and report these biases, and to use tools at our disposal to create fairer models, whether through preprocessing, in-processing, or postprocessing approaches. Equally important is that these fairness enhancements do not sacrifice the performance of the model to the detriment of its usefulness, and more importantly, that explanations are provided for the changes that occur during the fairness optimization. 
Examining the changes in feature importance as we optimize a model for fairness has the potential to enhance trust and willingness to apply fairness-optimized models to clinical applications.  Additionally, these explanations, when presented in a clinically accessible and relevant manner may provide valuable insights to assist healthcare providers in clinical decision-making and resource allocation.

\subsection{Strengths and Limitations}
Because we begin from a model optimized for predictive performance and then apply fairness regularization with performance constraints, our approach results in a fair model with minimal performance loss. 
A potential limitation of this study is that we use data from only one provider (HBFF), and therefore our findings may not be generalizable to all SUD treatment providers, facilities, or patients.  However, the HBFF dataset is a large multi-year multi-site dataset, and is therefore representative of multiple different regions across the US.  The sensitive attribute of race was used as entered in the EHR.  The use of race in observational studies has been shown to be problematic, for example where there may be conflicts between self-reported race versus provider-perceived \cite{polubriaginof_challenges_2019}.  Additionally, aggregating all non-Caucasian patients into one category due to the smaller sample sizes may lead to our findings not being generalizable to the individual race groups.

\section*{Acknowledgment}
This work was supported in part by the National Science Foundation under the Grants IIS-1741306 and IIS-2235548, and by the Department of Defense under the Grant DoD W91XWH-05-1-023.  This material is based upon work supported by (while serving at) the National Science Foundation.  Any opinions, findings, and conclusions or recommendations expressed in this material are those of the author(s) and do not necessarily reflect the views of the National Science Foundation. The authors would also like to thank the Hazelden Betty Ford Foundation for providing the data used in this study, and Dr. Ou Stella Liang for initial data extraction and cleaning.

\bibliographystyle{IEEEtran}

\bibliography{preprint}

\end{document}